\documentclass{article} 
\usepackage{iclr2017_conference,times}
\usepackage{hyperref}
\usepackage{url}
\usepackage[pdftex]{graphicx}
\usepackage[utf8]{inputenc} 
\usepackage[T1]{fontenc}    
\usepackage{booktabs}       
\usepackage{amsfonts}       
\usepackage{nicefrac}       
\usepackage{microtype}      
\usepackage{graphicx}
\usepackage{pifont}

\title{FreezeOut: Accelerate Training by Progressively Freezing Layers}

\author{Andrew Brock, Theodore Lim, \& J.M. Ritchie\\
School of Engineering and Physical Sciences\\
Heriot-Watt University\\
Edinburgh, UK \\
\texttt{\{ajb5, t.lim, j.m.ritchie\}@hw.ac.uk} \\
\And
Nick Weston \\
Renishaw plc \\
Research Ave, North \\
Edinburgh, UK \\
\texttt{Nick.Weston@renishaw.com} \\
}

\newcommand{\subf}[2]{%
  {\small\begin{tabular}[b]{@{}c@{}}
  #1\\#2
  \end{tabular}}%
}

\iclrfinalcopy 
\definecolor{mygreen}{rgb}{0.032, 0.6392, 0.2039}

\begin{document}

\maketitle

\begin{abstract}

The early layers of a deep neural net have the fewest parameters, but take up the most computation. In this extended abstract, we propose to only train the hidden layers for a set portion of the training run, freezing them out one-by-one and excluding them from the backward pass. Through experiments on CIFAR, we empirically demonstrate that FreezeOut yields savings of up to 20\% wall-clock time during training with 3\% loss in accuracy for DenseNets, a 20\% speedup without loss of accuracy for ResNets, and no improvement for VGG networks. Our code is publicly available at https://github.com/ajbrock/FreezeOut
\end{abstract}

\section{Introduction}

Layer-wise pre-training of neural nets \citep{GreedyLayer} \citep{StackedAuto} has largely been replaced by careful initialization strategies and fully end-to-end training. Successful techniques such as DropOut \citep{Dropout} and Stochastic Depth \citep{stdpth}, however, suggest that it is not necessary to have every unit in a network participate in the training process at every training step. Stochastic Depth leverages this to both regularize and reduce computational costs by dropping whole layers at a time (though it requires the use of residual connections \citep{PreAc}), while DropOut's use of masks does not by default result in a computational speedup.

In this work, we are concerned with reducing the time required for training a network by only training each layer for a set portion of the training schedule, progressively "freezing out" layers and excluding them from the backward pass. This technique is motivated by the observation that in many deep architectures, the early layers take up most of the budget, but have the fewest parameters and tend to converge to fairly simple configurations (e.g. edge detectors), suggesting that they do not require as much fine tuning as the later layers, where most of the parameters reside.

This work is also motivated by a (wholly unverified) hypothesis that the success of SGDR \citep{SGDR} is partially due to certain layers achieving near-optimality after every annealing-step before restart. We suspect that this results in faster convergence due to a reduction in internal covariate shift \citep{Bnorm}: if a unit's gradient is near zero, the unit will remain near its local optimum even when the learning rate is kicked back up on warm restart.

\section{FreezeOut}

\begin{figure}[htbp]
  \centering
  \begin{tabular}{cc}
  \subf{\includegraphics[scale=0.41]{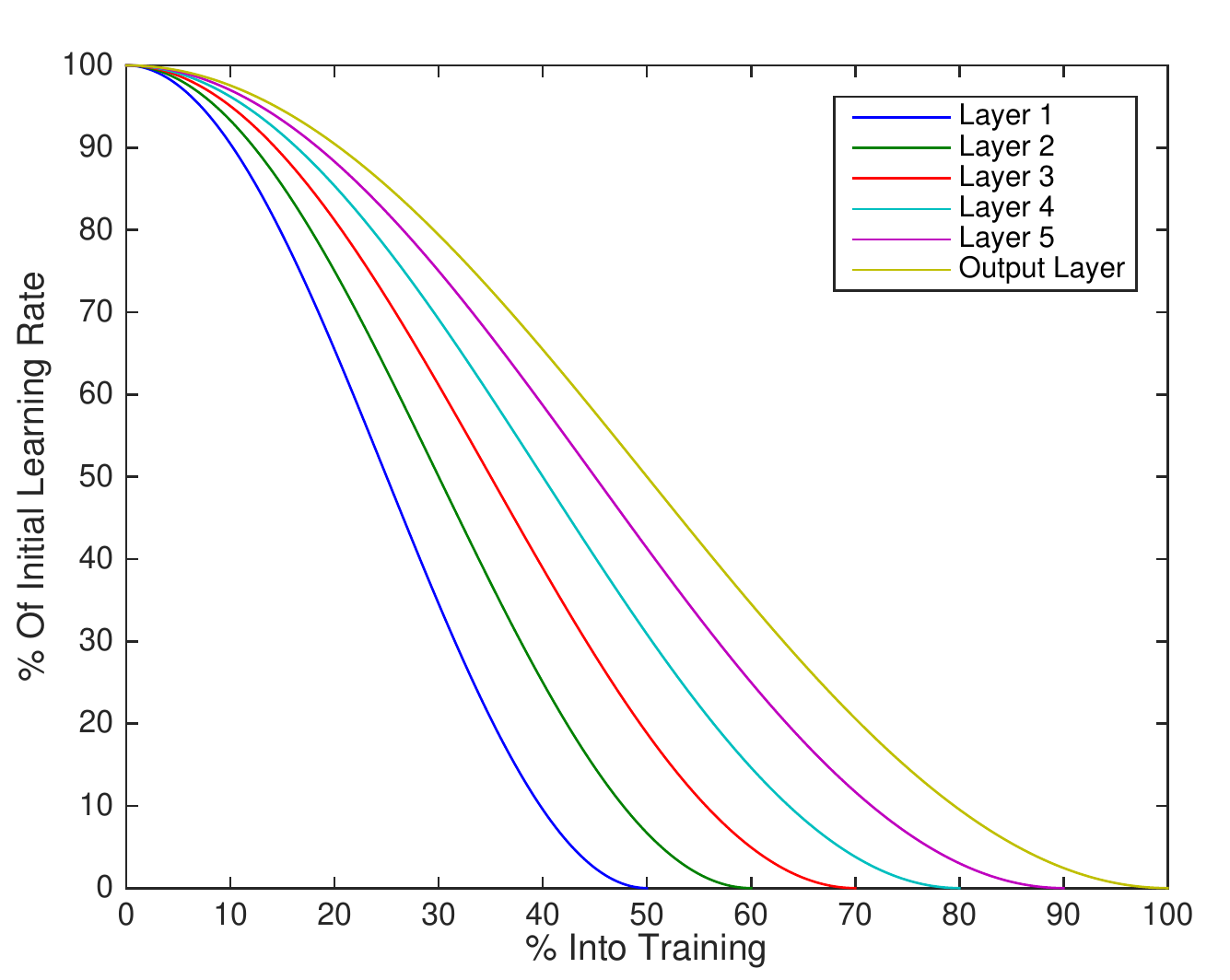}}{(a) Unscaled Linear Schedule}

&
 \subf{\includegraphics[scale=0.41]{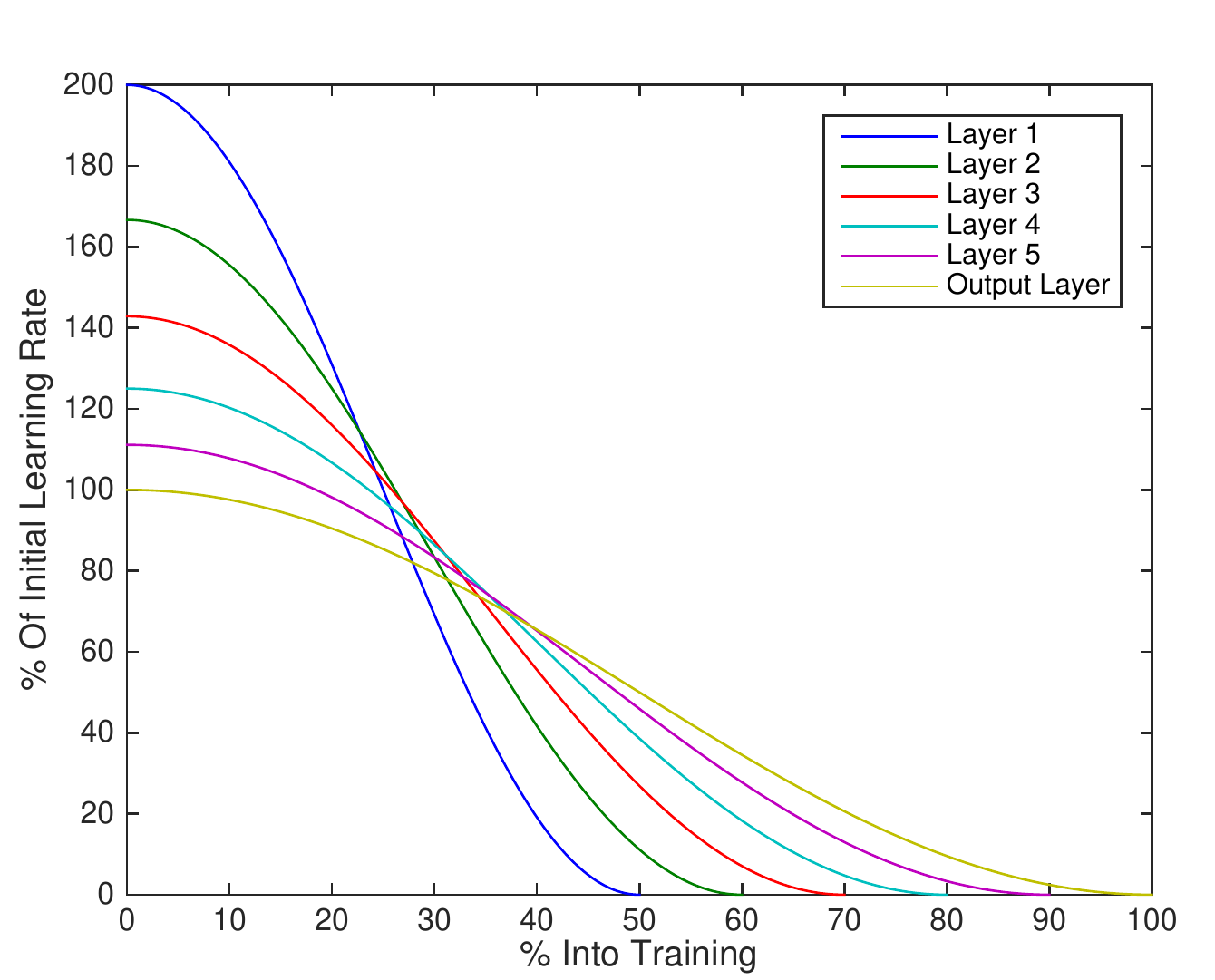}}{(b) Scaled Linear Schedule}
  \\
  \subf{\includegraphics[scale=0.41]{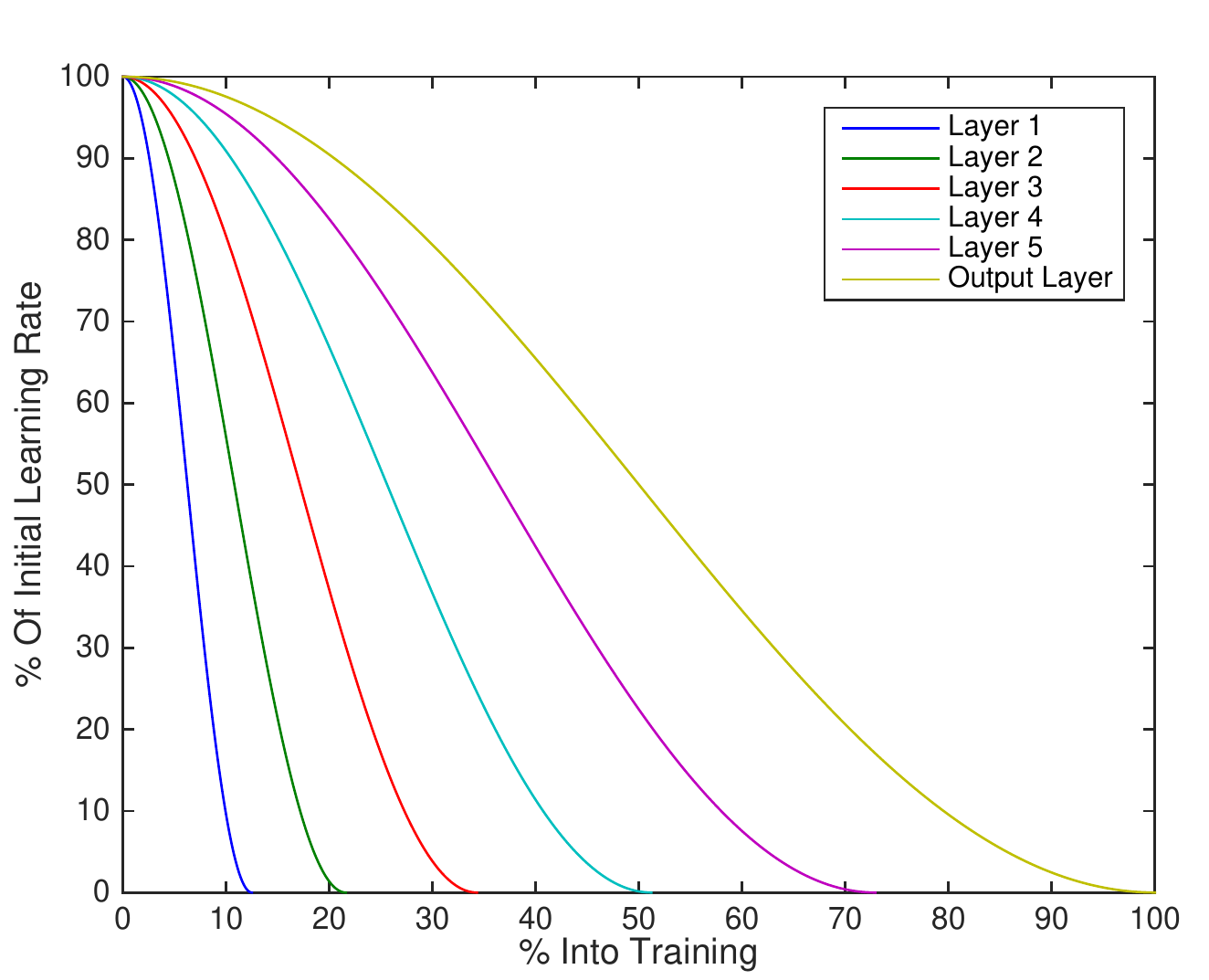}}{(c) Unscaled Cubic Schedule}

&
 \subf{\includegraphics[scale=0.41]{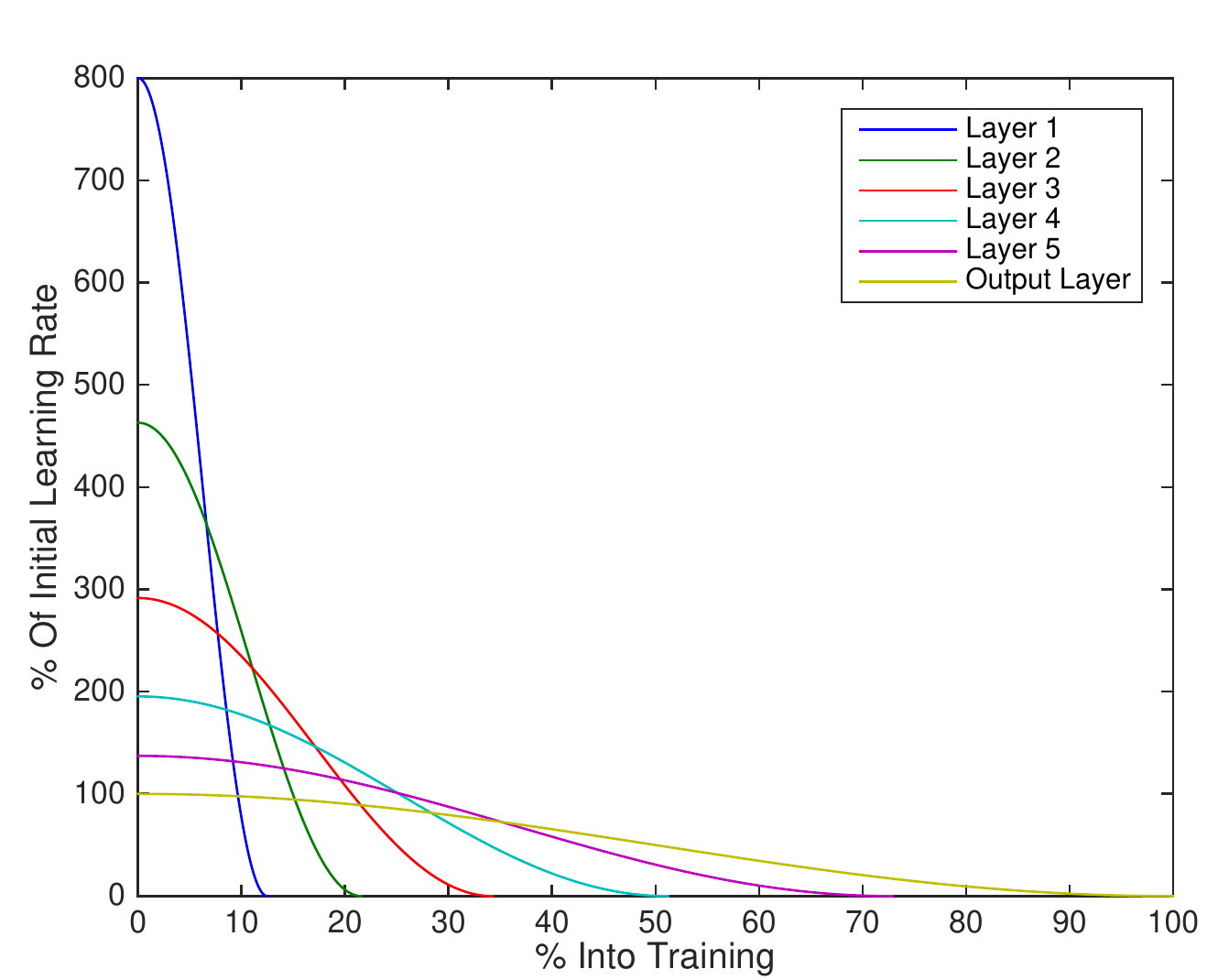}}{(d) Scaled Cubic Schedule}
 \end{tabular}
 \caption{Per-Layer Learning Rate Schedules for a 5-hidden-layer network with $t_0=0.5$.}
 \label{LR_Sched} 
\end{figure}

We propose a simple modification to the standard backprop+SGD pipeline to reduce training time. FreezeOut employs cosine annealing (as proposed by \citep{SGDR}) without restarts (as used by \citep{Shake}) with a layer-wise schedule, where the first layer's learning rate is reduced to zero partway through training (at $t_0$), and each subsequent layer's learning rate is annealed to zero some set time thereafter. Once a layer's learning rate reaches zero, we put it in inference mode and exclude it from all future backward passes, resulting in an immediate per-iteration speedup proportional to the computational cost of the layer.


In the simplest version of FreezeOut, each layer $L_i$ starts with a single fixed learning rate $\alpha$, which anneals to zero at $t_i$, where $t_i$ is linearly spaced between a user-selected $t_0$ and the total number of iterations, as shown in Figure ~\ref{LR_Sched}(a). Each layer's learning rate at iteration $t$ is thus given as
\begin{equation}
\alpha_i(t) = 0.5*\alpha_i(0) (1+cos(\pi t/t_i))
\end{equation}

 We experiment with varying two aspects of this strategy. First, we consider scaling the initial layer-wise learning rate to be $\alpha_i(0) = \alpha / t_i $, where $\alpha$ is the base learning rate (and the learning rate of the final layer). This scaling, shown in Figure ~\ref{LR_Sched} (b), causes each layer's learning curve to integrate to the same value, meaning that each layer travels the same distance in the weight space (modulo its observed gradients and weight dimension) despite the reduced number of steps taken.

Second, we vary the strategy for the how $t_0$ relates to the $t_i$s of the remaining layers. The simple version we explore takes the $t_i$ values determined by the linear scheduling rule, and cubes them with $t_{i(cubed)} = t_{i(linear)}^3$, as shown in Figure ~\ref{LR_Sched} (c) and (d). This gives more priority (in terms of training time) to later layers than does the linear schedule. As with the linear schedule, we consider an "unscaled" variant where the $\alpha_i$ values are identical, and a "scaled" variant where the $\alpha_i$ values are scaled based on the cubed $t_i$ values. Throughout this paper, we refer to $t_i$s with respect to their uncubed values, such that such that a user selected $t_0=0.5$ results in a cubed  $t_0=0.125$.

FreezeOut adds two user decisions--the choice of $t_0$ and the choice of FreezeOut strategy--to the standard choices of initial learning rate and number of training iterations. In the following section, we empirically investigate the relative merit of each of the four strategies, and provide recommendations for a default configuration, reducing the need for user-tuning of the hyperparameters. FreezeOut is easy to implement in a dynamic graph framework, requiring approximately 15 unique lines of code in PyTorch \citep{PyTorch}.


\section{Experiments}
\begin{figure}[tbp]
\begin{center}
{\includegraphics[width=0.92\linewidth]{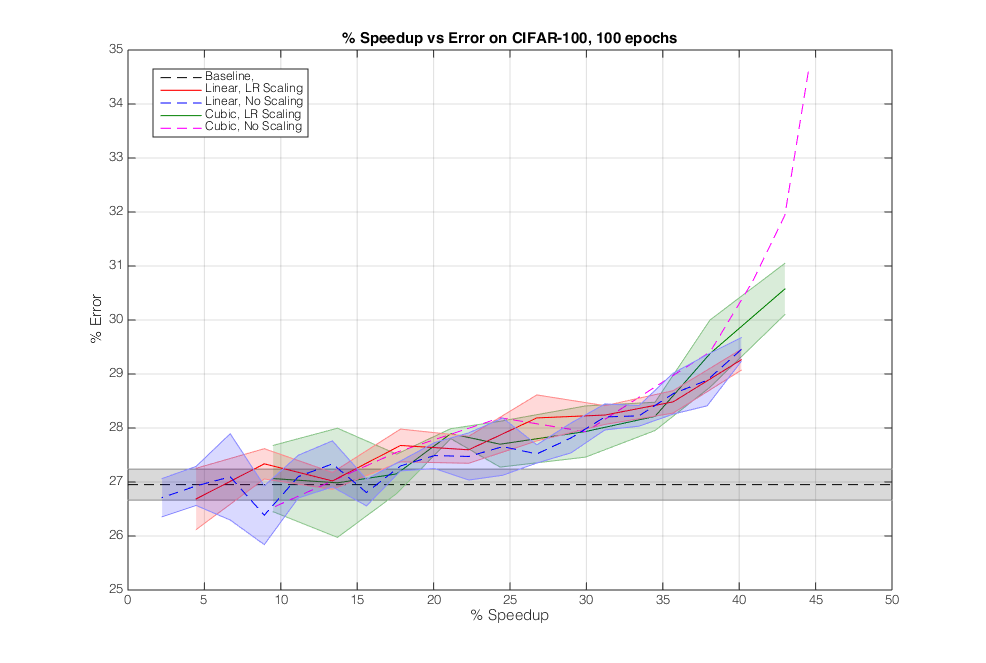}}
\end{center}
\caption{FreezeOut results for k=12, L=76 DenseNets on CIFAR-100 for 100 epochs. Shaded areas represent one standard deviation from the mean across 2-5 training runs.}
  \label{Results_plot_1}
\end{figure}

We modify publicly available implementations of DenseNets \citep{DenseNets}\footnote{https://github.com/bamos/densenet.pytorch}, Wide ResNets \citep{wide}\footnote{https://github.com/xternalz/WideResNet-pytorch}, and VGG \citep{VGG}  and test each of the four scheduling strategies across a wide range of $t_0$ values on CIFAR-100 and CIFAR-10. Our PyTorch \citep{PyTorch} code is publicly available.\footnote{https://github.com/ajbrock/FreezeOut}

\subsection{DenseNet Experiments}
Unless otherwise noted, all DenseNet experiments were performed using DenseNet-BC models with a growth rate of 12 and a depth of 76, and we swap the order of BatchNorm \citep{Bnorm} and ReLU from the original implementation (such that we perform ReLU-BatchNorm-Convolution). We make use of standard data augmentation, train using SGD with Nesterov Momentum \citep{Nesterov} and report results on the test set after training is completed. We compare achieved speedups (in terms of observed wall clock time relative to a non-FreezeOut baseline) to performance on the test set, and are primarily interested in the accuracy reduction incurred for a given reduction in training time.

For the speedups presented here, we use the values attained by running each test sweep in a controlled environment, on a GTX1080Ti with no other programs running. The actual speedups we observe in practice are slightly different, as we  run the experiments on shared servers, with various types of GPUs, and with many other programs also running. 

We found that a quick back-of the envelope calculation based on the reduction in operations accurately estimates the obtained speedups. Given $c_i$, the computational cost for the forward pass through each convolutional layer, and noting that the cost of a full forward-backward pass through that layer is $2c_i$, we calculate the baseline computational cost as the sum of each layer's cost: 

\begin{equation}
 C = \Sigma (2c_i \times n_{itr})
\end{equation}with $n_{itr}$ the total number of training iterations. The compute cost for training with FreezeOut is:  
\begin{equation}
C_f = \Sigma ((1+t_i) \times c_i \times n_{itr})
\end{equation}
and the approximate relative speedup is simply the ratio $1 - C_f/C$. We found that this accurately estimates speedup for cubic scheduling, but underestimates the speedups (e.g. the actual speedup is greater than the calculation gives) for linear scheduling; multiplying the predicted speedup by a correction factor of 1.3 resolves this almost entirely. This estimate neglects computational overhead and the cost of other operations (nonlinearities and BatchNorm), but this seems to be offset by the fact that a frozen layer's batchnorm costs are reduced (no longer having to calculate means and variances), and we find that this is generally a reliable estimate of the achieved speedups for a given setting.  

\begin{figure}[tbp]
\begin{center}
{\includegraphics[width=0.92\linewidth]{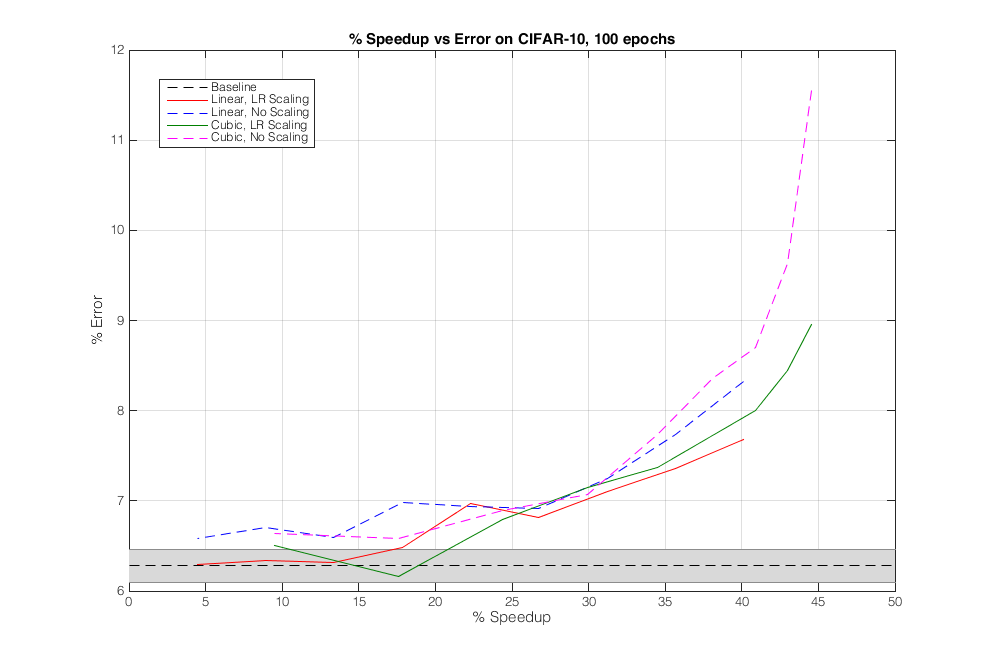}}
\end{center}
\caption{FreezeOut results for k=12, L=76 DenseNets on CIFAR-10 for 100 epochs.}
  \label{CIFAR10_fig}
\end{figure}

Our most-investigated setup is shown in Figure ~\ref{Results_plot_1}, where we train on CIFAR-100 for 100 epochs and repeat each experimental setting 2 to 5 times. We also train for a single pass on CIFAR-10 for 100 epochs (Figure ~\ref{CIFAR10_fig}). 

We observe a clear speedup versus accuracy tradeoff. For every strategy, we observe a speedup of up to 20\%, with a maximum relative 3\% increase in test error. Lower speedup levels perform better and occasionally outperform the baseline, though given the inherent level of non-determinism in training a network, we consider this margin insignificant.

%
%

Whether this tradeoff is acceptable is up to the user. If one is prototyping many different designs and simply wants to observe how they rank relative to one another, then employing higher levels of FreezeOut may be tenable. If, however, one has set one's network design and hyperparameters and simply wants to maximize performance on a test set, then a reduction in training time is likely of no value, and FreezeOut is not a desirable technique to use.

Based on these experiments, we recommend a default strategy of cubic scheduling with learning rate scaling, using a $t_0$ value of 0.8 before cubing (so $t_0 = 0.5120$) for maximizing speed while remaining within an envelope of 3\% relative error. As a close alternative, we suggest linear scheduling without learning rate scaling, using $t_0=0.5$. The user is, of course, free to select parameters to fit a particular point on the design curve.

\subsection{Wide ResNet experiments}
\begin{figure}[tbp]
\begin{center}
{\includegraphics[width=0.92\linewidth]{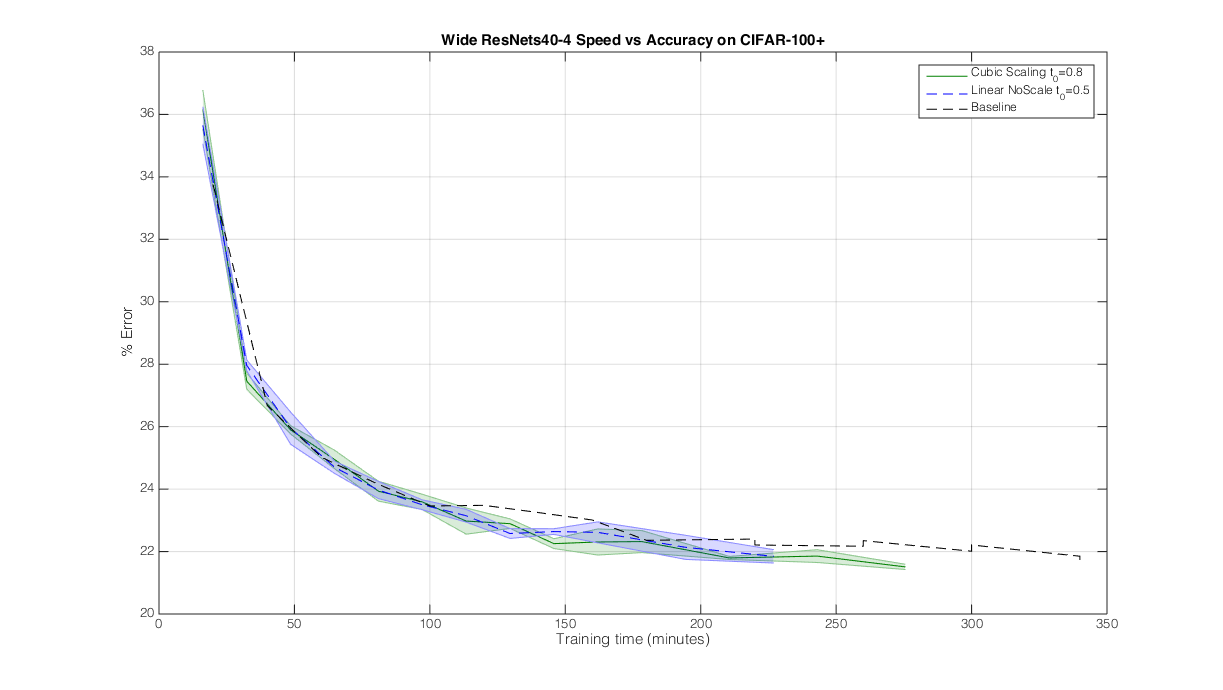}}
\end{center}
\caption{FreezeOut results for WRN40-4 on CIFAR-100. Shaded areas represent one standard deviation from the mean across 3 training runs for Cubic Scaled and 4 training runs for Linear Unscaled.}
  \label{WRN_fig}
\end{figure}
We next investigate the feasibility of FreezeOut for use in Residual architectures \citep{PreAc}. We train Wide ResNets \citep{wide} with depth of 40 and widening factor of 4, and compare our two recommended strategies against a no-FreezeOut baseline. We vary the number of epochs for which we train, and investigate how FreezeOut performs against a baseline of simply training for fewer epochs (e.g. a network trained with FreezeOut for 100 epochs has approximately the same training time as a network trained without FreezeOut for 80 epochs). The results of this investigation are shown in  Figure~\ref{WRN_fig}. FreezeOut appears to be better suited to Wide ResNets than DenseNets, achieving \textit{higher} accuracy even when trained for the same number of epochs (the final point on the Cubic Scaled curve and the final point on the Baseline curve were both trained for the same number of iterations).  

\subsection{VGG Experiments}
\begin{figure}[tbp]
\begin{center}
{\includegraphics[width=0.92\linewidth]{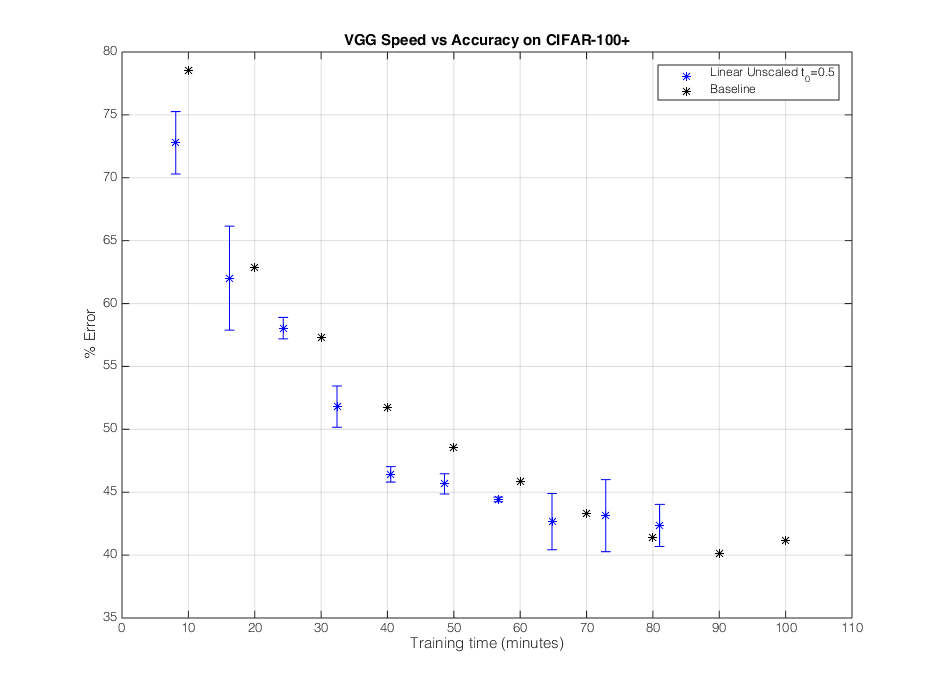}}
\end{center}
\caption{FreezeOut results for VGG-16 on CIFAR-100. Error Bars represent a single standard deviation from the mean across three training runs. Error bars instead of shaded error lines used here for improved clarity.}
  \label{VGG_fig}
\end{figure}
Finally, we investigate the use of FreezeOut on an architecture without skip connections. We employ a variant of the VGG-16 \citep{VGG} architecture with Batch Normalization \citep{Bnorm}, no Dropout \citep{Dropout}, and 512 units in each of the fully connected layers rather than 4096. We perform a more limited set of experiments using the Linear Unscaled strategy against a baseline without FreezeOut, as shown in Figure~\ref{VGG_fig}. FreezeOut appears to be less well-suited for use in VGG, suggesting that skip connections (either dense or residual) may be an important element enabling FreezeOut to work for the other architectures we investigate.

\section{Conclusion}

In this extended abstract, we presented FreezeOut, a simple technique to accelerate neural network training by progressively freezing hidden layers. 

\subsubsection*{Acknowledgments}

This research was made possible by grants and support from Renishaw plc and the Edinburgh Centre For Robotics. The work presented herein is also partially funded under the European H2020 Programme BEACONING project, Grant Agreement nr. 687676.

\bibliography{nips_2017}
\bibliographystyle{iclr2017_conference}

\end{document}